\documentclass[runningheads,a4paper]{llncs}

\usepackage{graphicx}
\usepackage{amssymb}
\usepackage{amsmath}
\usepackage{multirow}
\usepackage{rotating}
\usepackage{epsf} 
\usepackage{caption}
\usepackage{url}
\usepackage{booktabs}
\usepackage[rgb]{xcolor}
\usepackage{cite}
\usepackage{filecontents}

\urldef{\mailsa}\path|{lenka.skanderova,|
\urldef{\mailsb}\path|tomas.fabian}@vsb.cz| 
\newcommand{\keywords}[1]{\par\addvspace\baselineskip
\noindent\keywordname\enspace\ignorespaces#1}

\begin{document}

\mainmatter  

\title{Study of the Influence of the Number Normalization Scheme Used in Two Chaotic Pseudo Random Number Generators Used as the Source of Randomness in Differential Evolution}

\titlerunning{Study of the Influence of the Number Normalization Scheme Used in Two Chaotic Pseudo Random Number Generators Used as the Source of Randomness in Differential Evolution}

%
%
\author{Lenka Skanderova%
\thanks{The following grants are acknowledged for the financial support provided for this research: Grant Agency of the Czech Republic - GACR P103/13/08195S, is partially supported by Grant of SGS No. SP2014/42, VŠB - Technical University of Ostrava, Czech Republic, by the Development of human resources in research and development of latest soft computing methods and their application in practice project, reg. no. CZ.1.07/2.3.00/20.0072 funded by Operational Programme Education for Competitiveness.} \and Tomas Fabian}
\authorrunning{Study of the Influence of the Number Normalization Scheme Used in Two Chaotic Pseudo Random Number Generators Used as the Source of Randomness in Differential Evolution}

\institute{Department of Computer Science,\\ 
Faculty of Electrical Engineering and Computer Science,\\
 VSB - Technical University of Ostrava,\\
17.listopadu 15/2172, 70800 Ostrava, Czech Republic\\
\mailsa\\
\mailsb\\
\url{http://www.cs.vsb.cz}}

%
%

\toctitle{Study of the Influence of the Number Normalization Scheme Used in Two Chaotic Pseudo Random Number Generators Used as the Source of Randomness in Differential Evolution}
\maketitle

\begin{abstract}
In many publications, authors showed that chaotic pseudo random number generators (PRNGs) may improve performance of the evolutionary algorithms. In this paper, we use two chaotic maps Gingerbread man and Tinkerbell as the chaotic PRNGs instead of the classical PRNG in the differential evolution. Numbers generated by this maps are normalized to the unit interval by three different methods -- operation modulo, straightforward number normalization where we know minimal and maximal generated number and arctangent of the two variables $x$ and $y$, where numbers $x$ and $y$ are generated by the Gingerbread man map and Tinkerbell map. The first goal of this paper is to show whether the differential evolution convergence speed might be affected by the way how we normalize number generated by the chaotic map. The second goal is to find out the influence of the probability distribution function of the selected chaotic PRNGs. The results mentioned below showed that the selected normalization method may improve differential evolution convergence speed, especially in the case of arctangent and straightforward number normalization, where we know the minimal and maximal generated numbers. 

\end{abstract}

\section{Introduction}
\label{sec:intro}

Interconnection between chaos and randomness is known very long time. In 1940, J. von Neumann used logistic map as the chaotic pseudo random number generator (cPRNG). From this year, chaos as the PRNG has been used in various research areas: cryptography (\cite{crypt1}, \cite{crypt2}); image encryption (\cite{imencr1}, \cite{imencr2}); new PRNG research (\cite{newPRNG1}, \cite{newPRNG2}); evolutionary algorithms (EAs). In the last mentioned area, chaos have been used successfully as the chaotic PRNG (cPRNG) for example in the bee colony algorithm (\cite{ea1}), particle swarm optimization (PSO) (\cite{ea2}), genetic algorithm (\cite{ea3}) or in differential evolution (DE) (\cite{ea4}). R.~Caponetto et al. used logistic map as the cPRNG in all phases of EA, where the random number is needed (\cite{ea11}). G.~Zilong et al. described a novel immune EA, where logistic map is used to generate the chaos sequence (\cite{ea12}). B.~Liu et al. used logistic map to improve PSO (\cite{ea13}). The utilization of analytic programming for a synthesis of control law for selected discrete chaotic systems is described where authors used logistic map and then Henon map (\cite{ea14}). B.~Alatas introduced twelve chaos-embedded PSO methods, where eight chaotic maps have been analyzed in the benchmark functions (\cite{ea15}). The same author used logistic map in the chaotic harmony search algorithm. Senkerik et al. used DE for the evolutionary tuning of controller parameters for the stabilization of different chaotic systems, where the selected controlled discrete chaotic systems (Burgers map, Delayed logistic map and Lozi map) are used also as cPRNGs to drive the mutation and crossover process (\cite{senk1}). L.~dos Santos Coelho and V.~C.~Mariani described PID controller tuned by firefly algorithm using Tinkerbell map (\cite{tink1}).

The motivation of this paper is that in the most publications dealing with the chaos powered EA there is not made clear whether the improvement of the EAs convergence speed stems from the uniqueness of the sequence of the numbers generated by the cPRNGs or by the probability distribution function (PDF) of the selected cPRNGs. In the most publications, authors use only one way of the number normalization and we spared the comparison of the number normalization schemes.

In our work, we used two-dimensional maps -- Gingerbread man and Tinkerbell as the cPRNGs in DE. Numbers generated by cPRNG are used in all cases where the randomness is needed in DE. As Gingerbread man and Tinkerbell maps might generate numbers outside of the unit interval, we have selected three schemes of number normalization. Operator modulo, normalization by bounds are traditional representatives of common normalization schemes. We have defined the third scheme in addition to overcome some peculiarities related with the preceding two schemes (eg. we need to know the bounds, modulo leads to over utilization of some sub-intervals of the generator). Normalized number is then used in DE and the convergence speed of DE to the global minimum is observed. 

The rest of the paper is organized as follows: The differential evolution is described briefly in the Section \ref{sec:DE}. In the Section \ref{sec:chaos}, the selected cPRNGs are introduced. The selected testing problems are mentioned in the Section \ref{sec:functions}. In the Section \ref{sec:met}, we are presenting our methods of number normalization. The methods of the analysis and setting of the DE are mentioned in the Section \ref{sec:set}. In the Section \ref{sec:res}, the results of the experiments are recorded and in the Section \ref{sec:concl}, the experiments results are discussed.

\section{Differential evolution}
\label{sec:DE}

DE belongs to the family of the evolutionary algorithms (EAs) working with the population of the individuals (\cite{anealing1, DE, DE2}). 

Here, let us describe the DE informally. The first population is generated randomly in the space of possible solutions. Then for each individual three random individuals (parents) are chosen. From these parents, we create a noise vector $\mathbf{v}$ according to the following equation

\begin{equation}
v_j = x_{r_3,j}^{G} + F(x_{r_1,j}^{G} - x_{r_2,j}^G) \,,
\label{eq:2}
\end{equation}

\noindent where $v_j$ denotes  j-th parameter of the noise vector, $x_{r_3}^{G}$ is the third randomly selected parent, $x_{r_1}^{G}$ is the first randomly selected parent and $x_{r_2}^G$ is the second randomly selected parent. The superscript $G$ means the actual generation, $F$ denotes mutation constant.

Then random number $r$ from the unit interval is generated for each parameter of the actual individual. If $r<CR$, where $CR$ is crossover probability, parameter from the noise vector is added to the trial individual, otherwise parameter from the actual individual is chosen. Now, the fitness value of the trial individual is computed. If it is better than fitness of the actual individual, the trial individual will be added to the new population, otherwise actual individual will be added. Process described above is repeated until some criterion of convergence is reached (\cite{DE}).

Beside the first variant of DE characterized by the Eq.~\eqref{eq:2} we have also included the variant DE/best/1/bin. The noise vector creation is described by the following equation  

\begin{equation}
v_j=x_{\text{best},j}^{G} + F(x_{r_2,j}^{G} - x_{r_3,j}^{G})\,,
\label{eq:DE/best/1/bin}
\end{equation}
 
\noindent where $x_{\text{best}}^{G}$ denotes the individual with the best fitness value in the actual generation (\cite{zelinka2}).

\section{Chaotic maps}
\label{sec:chaos}

This section contains description of the selected chaotic maps used as the cPRNGs in DE. We have selected Gingerbread man and Tinkerbell which might generate numbers outside the unit interval. In addition, they give promising results from the view of DE convergence speed and they are easy to implement.

\subsection{Gingerbread man map}
The Gingerbread man map (see Fig. \ref{fig:GingerBreadMan}) is a chaotic two-dimensional map which was studied by R.~Devaney (\cite{gbm}) since 1984 and is given by the following equation 

\begin{equation}
\begin{aligned}
x_{n+1} &= 1 - y_n + |x_n| \,,\\
y_{n+1} &= x_n \,.
\end{aligned}
\label{eq:ginger}
\end{equation}

In this paper, the initial values of $x$ and $y$ have been experimentally set to $x_0=9.0$ and $y_0=3.7$.

\subsection{Tinkerbell map}

\begin{figure}[ht]
\begin{minipage}[b]{0.45\linewidth}
\centering
\includegraphics[width=\textwidth]{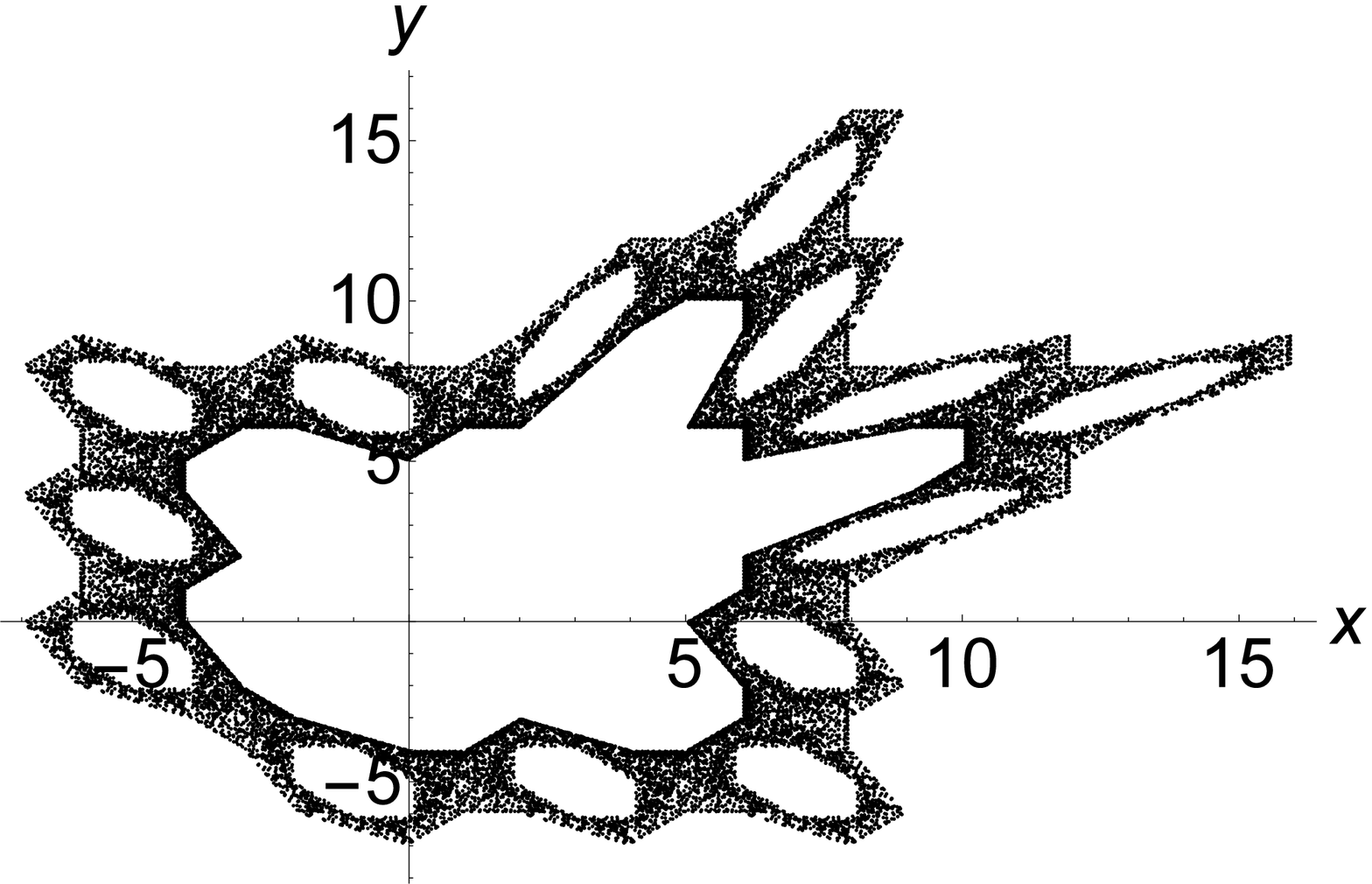}
	\caption{Gingerbread man map}
	\label{fig:GingerBreadMan}
\end{minipage}
\hspace{0.5cm}
\begin{minipage}[b]{0.45\linewidth}
\centering
\includegraphics[width=\textwidth]{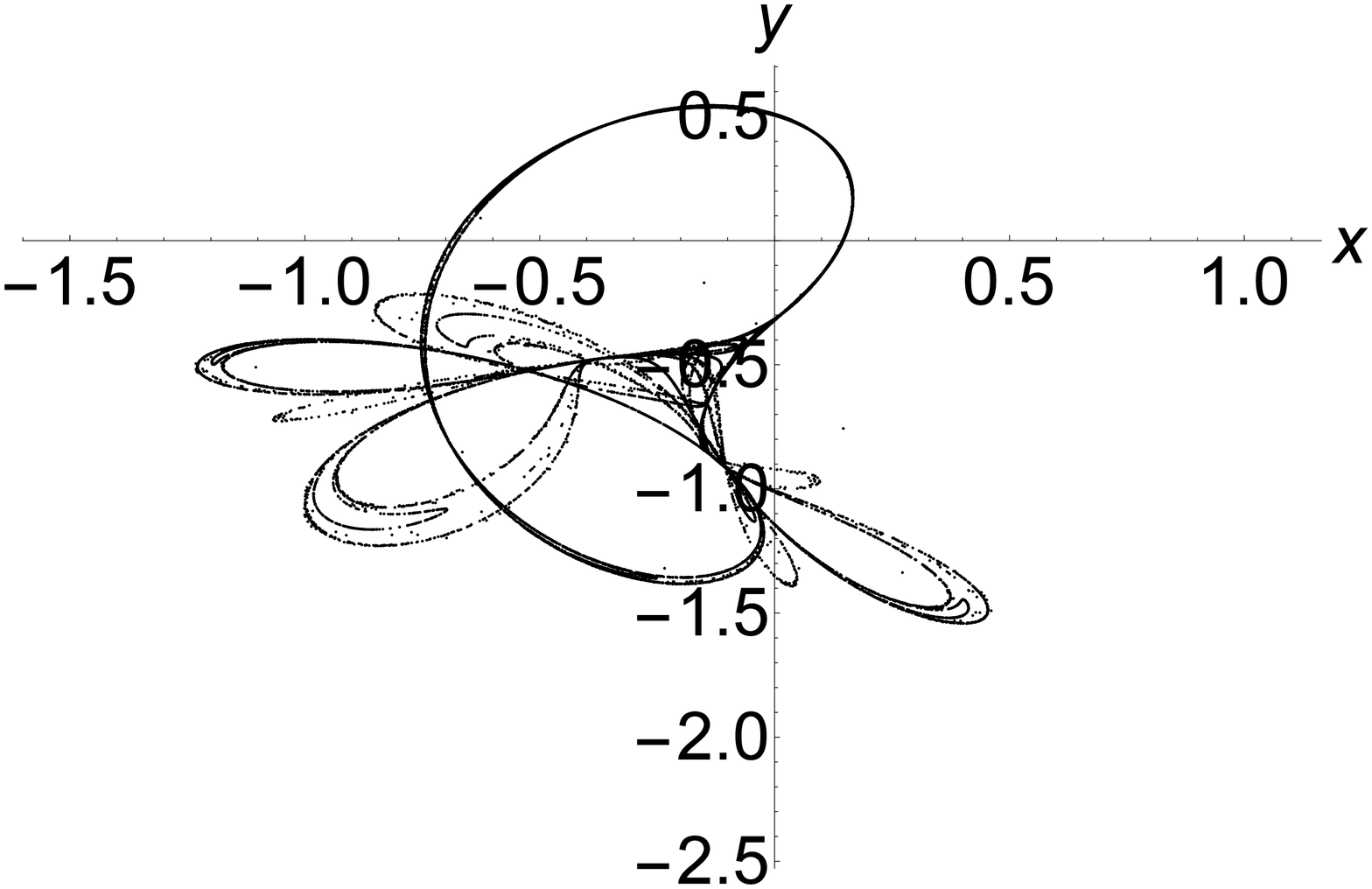}
	\caption{Tinkerbell map}
	\label{fig:Tinkerbell}
\end{minipage}
\end{figure}

Tinkerbell map (see the Fig. \ref{fig:Tinkerbell}) is the strange attractor with a fractal basin boundary and it was proposed by H.~E.~Nusse and J.~A.~Yorke (\cite{tinkerbell0}). It is given by the following equation

\begin{equation}
\begin{aligned}
x_{n+1} &= x_{n}^2 - y_{n}^2 + ax_n + by_n\,, \\
y_{n+1} &= 2x_n y_n + cx_n + dy_n\,.
\end{aligned}
\label{eq:tinkerbell}
\end{equation}

In this paper the parameters have been set to $a=0.9$, $b=-0.6013$, $c=2.0$, and $d=0.5$ as suggested in (\cite{tinkerbell0}). The initial value of $x$ and $y$ have been experimentally set to $x_0=0.1$ and $y_0=-0.1$.

\section{Normalization of the number generated by the cPRNG}
\label{sec:met}

We have selected two discrete dynamical systems -- Gingerbread man and Tinkerbell to be used as the cPRNGs in DE. Because these cPRNGs might generate numbers outside the unit interval it is necessary to normalize that number to the unit interval. In our work, we have chosen operation modulo (Modulo), straightforward number normalization where we know minimal and maximal generated number (Bounds) and two-argument variant of arctangent (Atan2) where the real numbers $x$ and $y$ are generated by the Gingerbread man and Tinkerbell maps. Subsequently, we have modified the uniform PDF of Mersenne Twister (MT) to approximate the PDF of our cPRNGs. In the following paragraphs, we would like to clarify the reasons why we have chosen these three ways of number normalization.

As the first way, operation \textit{modulo} has been chosen. It is the easiest way how to normalize the numbers lying outside of the unit interval. Each number generated by the cPRNG is modified according to the following equation

\begin{equation}
z_{i} = |n_i| \mod 1\,,
\label{eq:modulo}
\end{equation}

\noindent where $n_i$ is number generated by the selected cPRNG, $mod$ denotes operator modulo and $z_i$ is the i-th normalized number. This way of number normalization has been successfully  used for example by D. D. Davendra et al. (\cite{senkmodulo}), where authors use Tinkerbell and others as the cPRNGs in a scatter search algorithm. The main problem of this scheme is its PDF, because different numbers generated by cPRNGs might be normalized to the same values. For example the sequence $\{1.2, 2.2, 3.2, \ldots\}$ will be normalized to the single value $0.2$.

The second way how to normalize the number generated by the cPRNG to the unit interval is the straightforward number normalization where we know minimum and maximum generated by the number generator (Bounds). This normalization scheme is given by the following equation

\begin{equation}
z_i = \frac{x_i - \min(x)}{\max(x) - \min(x)}\,,
\label{eq:normalization}
\end{equation}

\noindent where $x = (x_1, \ldots, x_n)$ and $z_i$ is the i-th normalized number. This scheme of normalization has been successfully used by L. dos Santos et al. (\cite{tink1, tink2}). The main problem of this way was that we do not know the minimal and maximal number generated by the Gingerbread man and Tinkerbell map. Due this bottleneck, for each chaotic map we had generated one billion numbers with accuracy to one hundred decimal places and minimal and maximal values were obtained.

The last scheme is the arctangent $\text{atan2}(y, x)$, which has been experimentally added. The main advantage is that there is not distortion of the PDF like in the case of \textit{modulo}. Function $\text{atan2}(y, x)$ computes the angle in the sampling plane corresponding to the phase angle of the point $(x, y)$. The number generated by the chaotic map is then modified according to the following equation

\begin{equation}
z_{i} = \text{atan2}(y_i - \overline{y},x_i - \overline{x})\,,
\label{eq:atan2}
\end{equation}

\noindent where $z_i$ denotes the i-th normalized number, $y_i$ is the $y$ coordinate of the selected chaotic map, $x_i$ is the $x$ coordinate of the selected chaotic map and ($\overline{x}$, $\overline{y}$) denotes the coordinates of the attractor center equaling to the center computed as the average of the samples generated in the preceding step.

\begin{figure}[h]
	\centering
	\begin{minipage}{0.45\linewidth}
		\includegraphics[width=\textwidth]{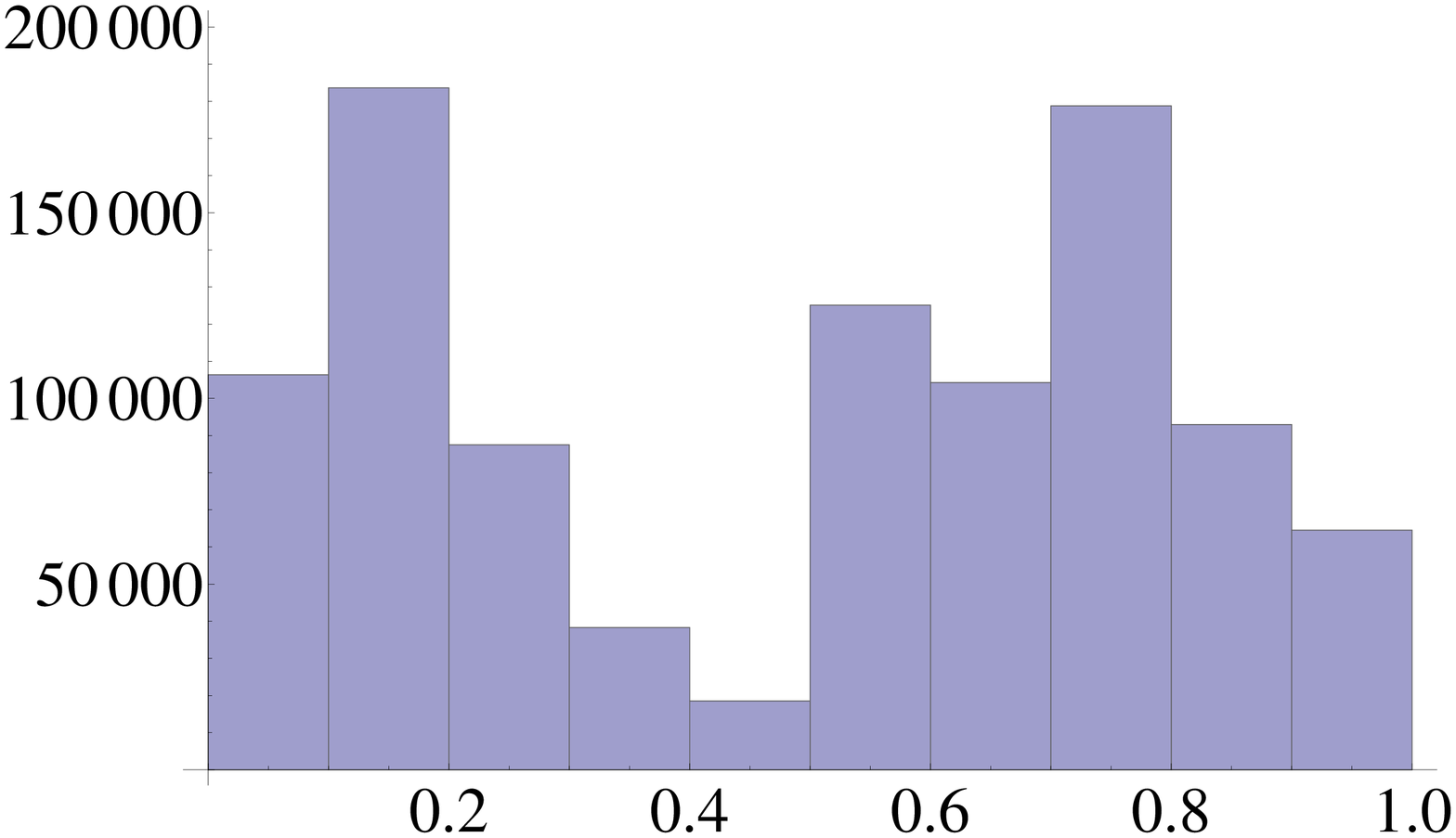}
		\caption{Histogram of the 1e+6 samples generated by the cPRNG using Tinkerbell and Atan2 normalizer}
	\label{fig:GBMAtan}
	\end{minipage}
	\hspace{0.5cm}
	\begin{minipage}{0.45\linewidth}
		\includegraphics[width=\textwidth]{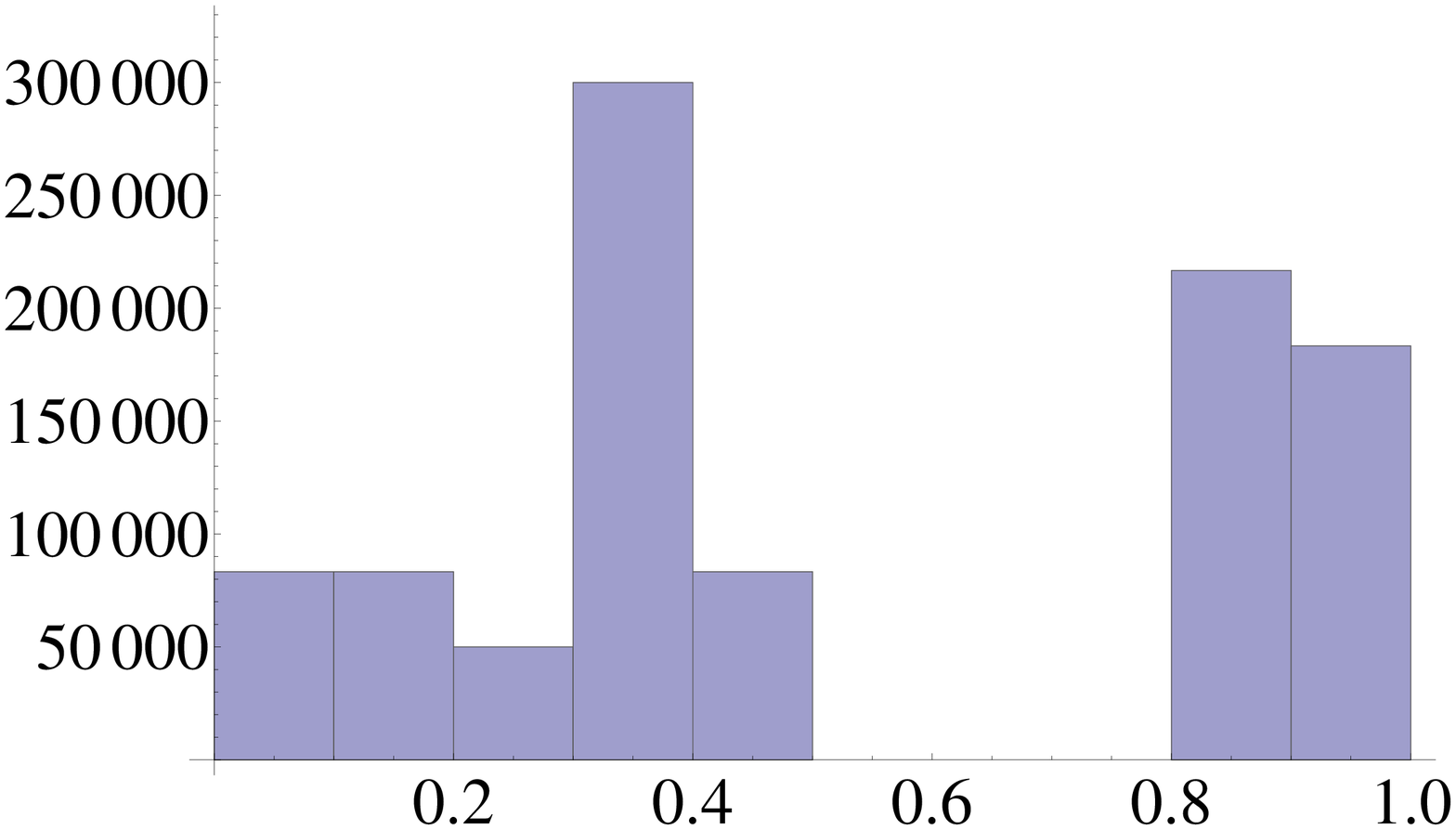}
		\caption{Histogram of the 1e+6 samples generated by the cPRNG using Gingerbread man and Atan2 normalizer}
	\label{fig:GBMAtan}
\end{minipage}\\
	\vspace{0.5cm}
	\begin{minipage}{0.45\linewidth}
		\includegraphics[width=\textwidth]{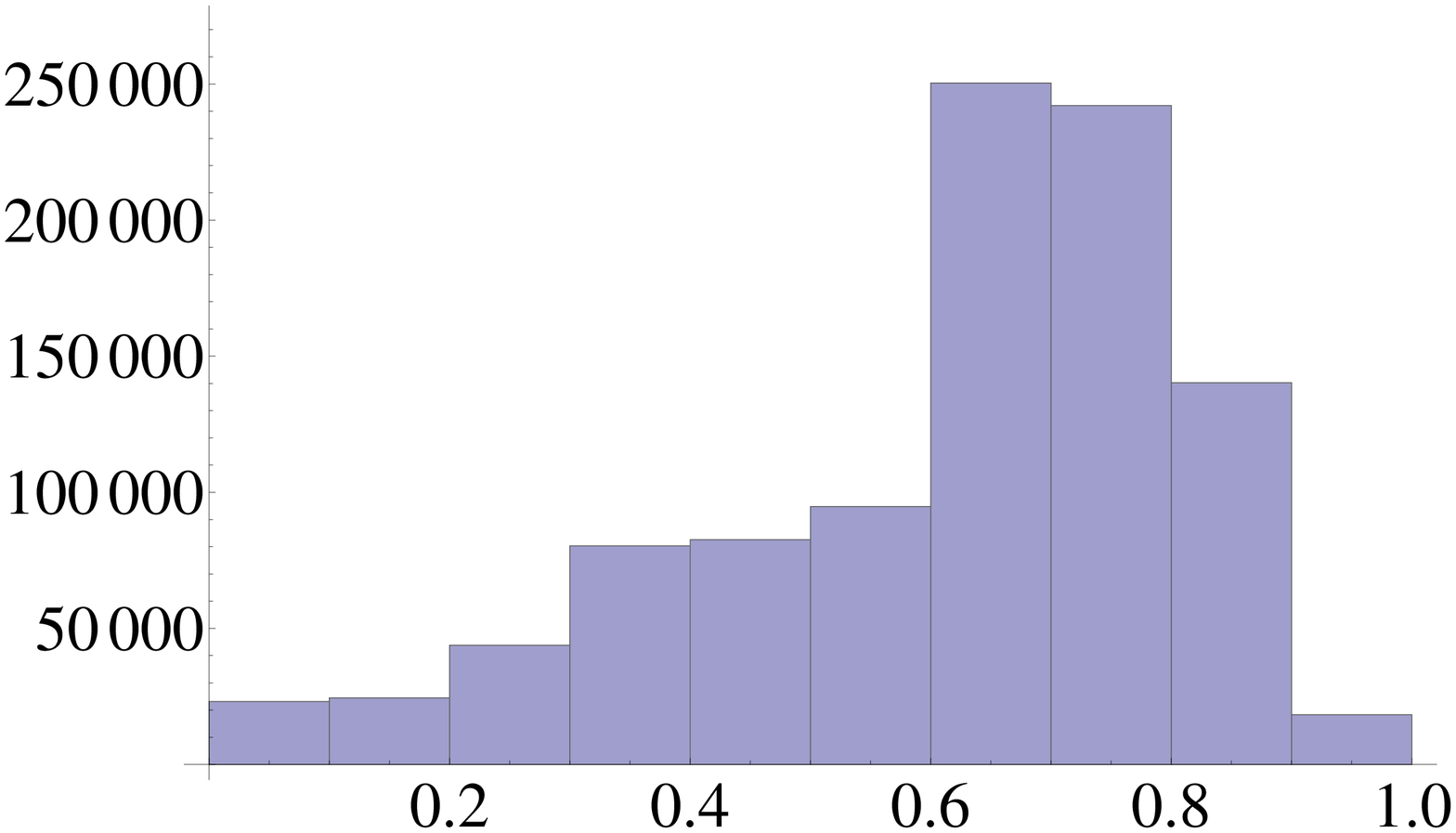}
		\caption{Histogram of the 1e+6 samples generated by the cPRNG using Tinkerbell and Bounds normalizer}
	\label{fig:GBMBounds}
	\end{minipage}
	\hspace{0.5cm}
	\begin{minipage}{0.45\linewidth}
		\includegraphics[width=\textwidth]{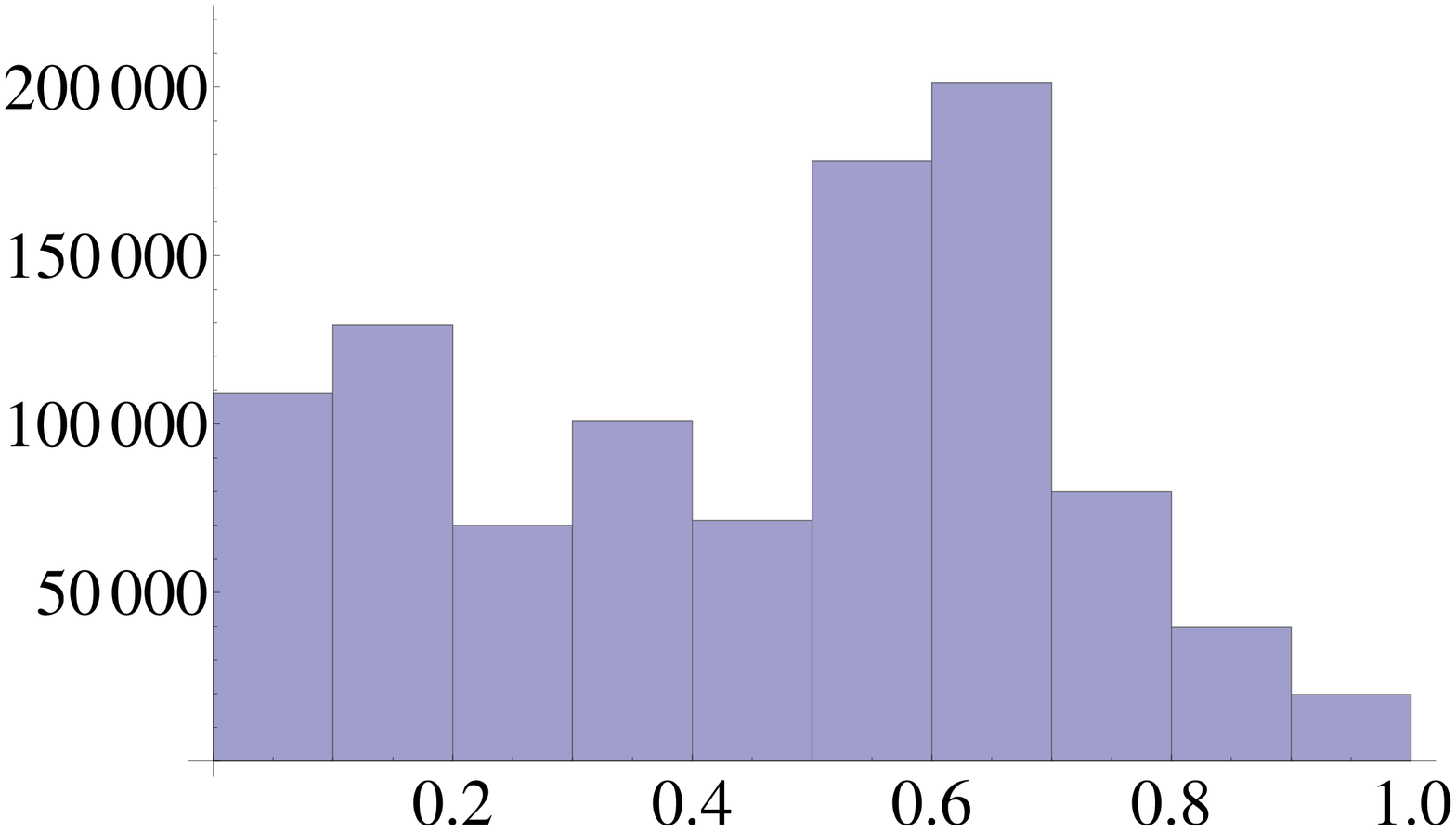}
		\caption{Histogram of the 1e+6 samples generated by of the cPRNG using Gingerbread man and Bounds normalizer}
	\label{fig:GBMBounds}
\end{minipage}\\
\vspace{0.5cm}
	\begin{minipage}{0.45\linewidth}
		\includegraphics[width=\textwidth]{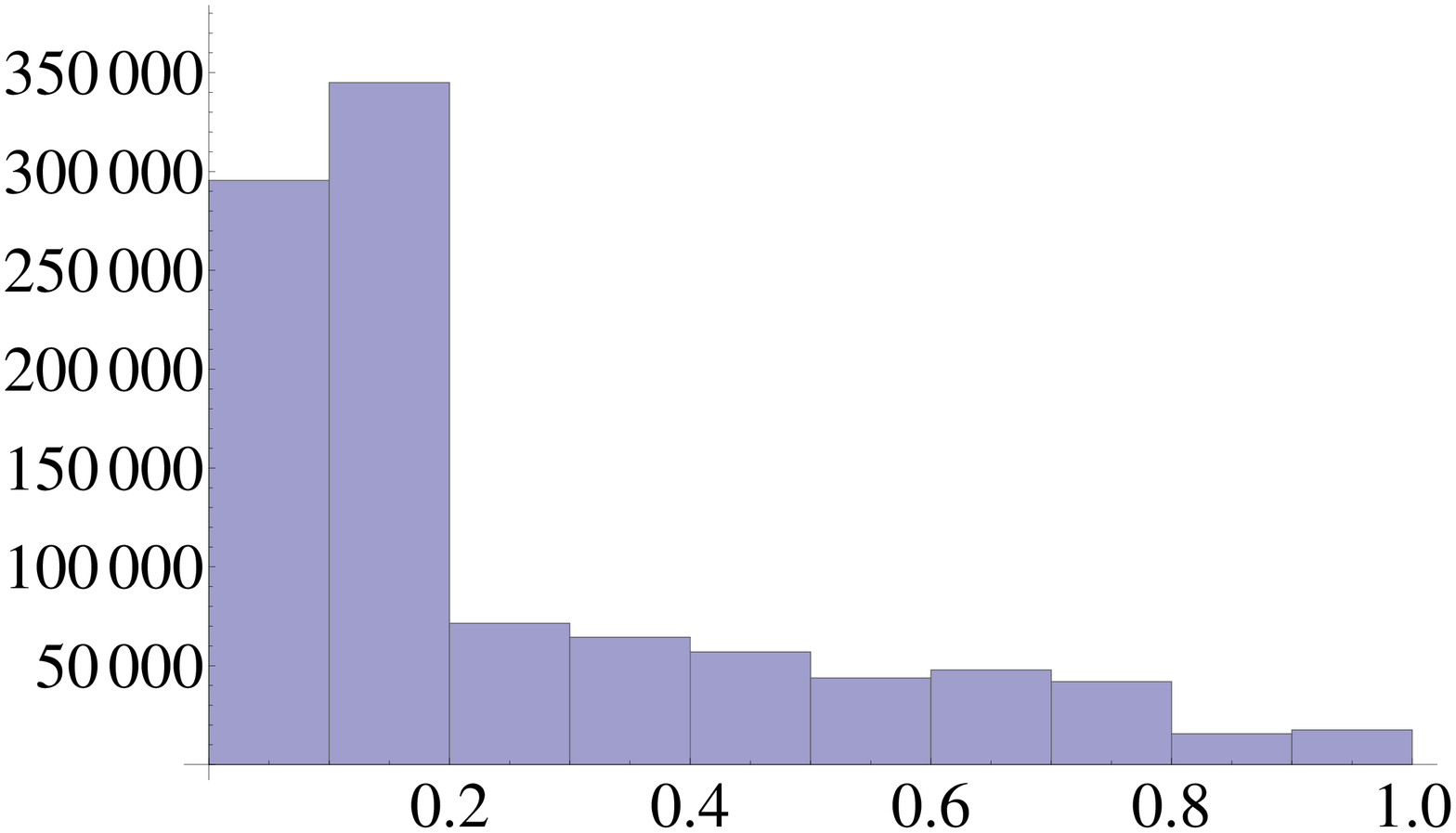}
		\caption{Histogram of the 1e+6 samples generated by the cPRNG using Tinkerbell and Modulo normalizer}
	\label{fig:GBMModulo}
\end{minipage}
\hspace{0.5cm}
\begin{minipage}{0.45\linewidth}
		\includegraphics[width=\textwidth]{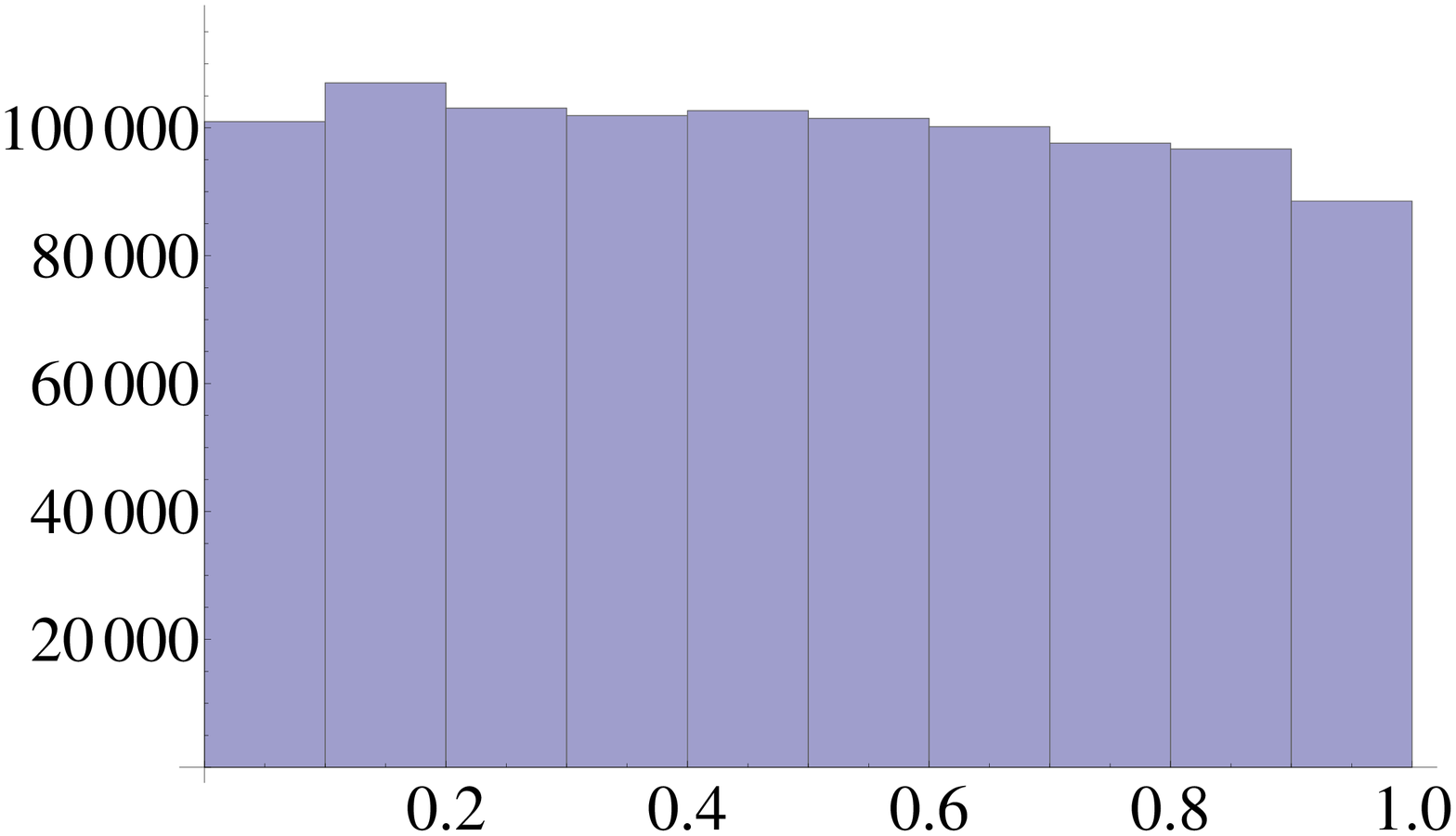}
		\caption{Histogram of the 1e+6 samples generated by the cPRNG using Gingerbread man and Modulo normalizer}
	\label{fig:GBMModulo}
\end{minipage}
\end{figure}

\section{Selected testing problems}
\label{sec:functions}

We have selected nine functions from the CEC2013 benchmark (\cite{cec2013}). For the specification of the selected function, please see (\cite{cec2013}). The reason of this choice is that CEC2013 benchmark provides 28 difficult functions and in the future we would like to extend our work to other functions from this benchmark and from the CEC2014 benchmark, see (\cite{cec2014}). 

\begin{table}[htb]
\renewcommand{\arraystretch}{1.2}
	\centering
	\caption{Selected functions from benchmark CEC2013}
		\begin{tabular}{l l l r}
		\hline
		\textbf{Cathegory} & \textbf{Funct.} & \textbf{Name} & \textbf{Global min.}\\
		\hline
		 \multirow{2}{*} {Unimodal} & $f_{1}$ & Sphere & -1400\\ 
																& $f_{5}$ & Different Powers & -1000\\
		\hline
		 \multirow{5}{*}{Basic multiomodal} & $f_{9}$ & Rotated Weierstrass & -600\\
																				& $f_{13}$ & Non-Continuous Rotated Rastrigin’s & -200\\
																				& $f_{15}$ & Rotated Schwefel's & 100\\
																				& $f_{16}$ & Rotated Katsuura & 200\\
																				& $f_{17}$ & Lunacek Bi Rastrigin & 300\\																		
		\hline
		\multirow{2}{*}{Composition} & $f_{22}$ & Composition function 2 &  800\\
																 & $f_{23}$ & Composition function 3 &  900\\		
		\hline
		\end{tabular}
	\label{tab:funct}
\end{table}

\section{Methods and experiment settings}
\label{sec:set}

In the first experiment, we have used Gingerbread man and Tinkerbell maps as the cPRNGs with the different number normalization schemes in DE/best/1/bin and DE/rand/1/bin. As the testing problems nine functions from CEC2013 have been chosen. For each function, three categories have been created according to the number normalization scheme described above and denoted as \textbf{Atan2}, \textbf{Bounds} and \textbf{Modulo}. Each experiment has been repeated fifty times. The results are reported in Tables \ref{tab:DEBest1BinD10T} -- \ref{tab:DErand1BinD30G}. The results represents the relative number of winnings of the given normalization scheme used in the cPRNG from the view of DE convergence speed. When two number normalization schemes reach the best results in the same time, they are recorded both as the best.


\begin{table}[h!]
\renewcommand{\arraystretch}{1.2}
	\centering
	\caption{Table of the selected symbols using in the text}
	\begin{tabular}{l@{\hskip 0.3in}l}
		\hline
		\textbf{Symbol} & \textbf{Meaning}\\
		\hline
		tMT & MT with the modified PDF according to the cPRNG using\\
				&	Tinkerbell.\\
		tMT, Atan2 & MT with the modified PDF according to the cPRNG\\ 
							 & using Tinkerbell and number normalization Atan2.\\
		tMT, Bounds & MT with the modified PDF according to the cPRNG \\
								& using Tinkerbell and number normalization Bounds.\\
		tMT, Modulo & MT with the modified PDF according to the cPRNG\\
								& using Tinkerbell and number normalization Modulo.\\
		gMT & MT with the modified PDF according to the cPRNG \\
				& using Gingerbread man.\\
		gMT, Atan2 & MT with the modified PDF according to the cPRNG\\
							 &	using Gingerbread man and number normalization Atan2.\\
		gMT, Bounds & MT with the modified PDF according to the cPRNG\\
								& using Gingerbread man and number normalization Bounds.\\
		gMT, Modulo & MT with the modified PDF according to the cPRNG \\
								& using Gingerbread man and number normalization Modulo.\\
		\hline
		$\mu_{A}$ & The mean of results of the DE using cPRNG\\ 
							& using number normalization Atan2.\\
		$\mu_{B}$ & The mean of results of the DE using cPRNG\\
							& using number normalization Bounds.\\
		$\mu_{M}$ & The mean of results of the DE using cPRNG\\
							& using number normalization Modulo.\\
		\hline
		$\mu_{A}^{tMT}$ & The mean of results of the DE using MT\\
										&	with the modified PDF according to the cPRNG\\
										&	using Tinkerbell and number normalization Atan2.\\
		$\mu_{B}^{tMT}$ & The mean of results of the DE using MT\\
										&	with the modified PDF according to the cPRNG\\
										& using Tinkerbell and number normalization Bounds.\\
		$\mu_{M}^{tMT}$ & The mean of results of the DE using MT\\
										& with the modified PDFaccording to the cPRNG\\
										& using Tinkerbell and number normalization Modulo.\\
		$\mu_{A}^{gMT}$ & The mean of results of the DE using MT\\ 
										& with the modified PDF according to the cPRNG\\
										& using Gingerbread man and number normalization Atan2.\\
		$\mu_{B}^{gMT}$ & The mean of results of the DE using MT\\
										& with the modified PDF according to the cPRNG\\
										& using Gingerbread man and number normalization Bounds.\\
		$\mu_{M}^{gMT}$ & The mean of results of the DE using MT\\
										& with the modified PDF according to the cPRNG\\
										& using Gingerbread man and number normalization Modulo.\\
		\hline
	\end{tabular}
	\label{tab:symb}
\end{table}

Now we would like to make clear motivation of the following experiments. The first goal was to find out which normalization scheme is the most successful from the view of DE convergence speed and the second goal was to investigate the influence of the PDF of the selected chaotic PRNGs. The results of the experiments mentioned in the Tables \ref{tab:DEBest1BinD10T} -- \ref{tab:DErand1BinD30G} (odd columns) give us the relative number of winnings of the normalization scheme using in the selected cPRNG. In our opinion, the greatest mean of the results of the cPRNG corresponds to the most successful normalization scheme and the PDF of the cPRNG fundamentally influences DE convergence speed. We have selected the following statistical methods to find out the greatest mean of the normalization schemes (columns of the mentioned Tables):

\begin{enumerate}
	\item It was necessary to find out if the results mentioned in the Tables  \ref{tab:DEBest1BinD10T} -- \ref{tab:DErand1BinD30G} are normally distributed. We have used Kologorov-Smirnov test for each column of these results. 
	\item When the normality is verified we have to find out if the variances of the columns of the mentioned tables can be considered as equal. 
	\item If the variances can be considered as equal, statistical test ANOVA will be used to find out if the means of columns can be considered as equal. 
	\item If the variances can not be considered as equal, we can use one-sided and two-sided T-tests with nonequivalent variances to find out which mean is the greatest.
	\item When the means can not be considered as equal the statistical one-sided T-tests with equivalent variances can be applied to find out which mean is the greatest. 
\end{enumerate}

When the greatest means are found we can compare the means of the columns of the cPRNGs with the columns of MT with the modified PDF according to these cPRNGs. If the normalization scheme fundamentally affects DE convergence speed, the results of the cPRNG and MT with the modified PDF according to this cPRNG will be comparable.

We have assumed that the results of the different number normalization schemes will be different. From this reason it was necessary to formulate the null $H_{0}$ and alternative hypothesis $H_{A}$ and the level of significance $\alpha$:

\begin{itemize}
	\item $H_{0}$: The means of results of the categories denoted as Atan2, Bounds and Modulo are \textbf{different}.
	\item $H_{A}$: The means of results of the categories denoted as Atan2, Bounds and Modulo are \textbf{the same}.
	\item The significance level has been chosen to be $\alpha=0.1$ (10\%).
\end{itemize}

Firstly it was necessary to verify whether the results mentioned in Tables \ref{tab:DEBest1BinD10T} -- \ref{tab:DErand1BinD30G} (odd columns) are normally distributed. The Kolmogorov-Smirnov test has been applied and normal distribution has been confirmed. Then we have performed the tests of variance equality, where Atan2, Bounds and Modulo have been compared with each other. It was found that we can consider the variances of the categories as equal in the three of four cases. In the case of DE/rand/1/bin powered by cPRNG using Gingerbread man the variances of Atan2, Bounds and Modulo can not be considered as equal. To compare means of the remaining data sets statistical method ANOVA has been applied. The results are mentioned in the Table \ref{tab:Anova1}. We can see that in the case of DE/best/1/bin powered by cPRNG using Gingerbread man map means can be considered as equal. That means the success of all three normalization schemes is comparable.
 
Statistical one-sided (if it was necessary two-sided) T-test has been performed for data sets DE/best/1/bin powered by the cPRNG using Tinkerbell, and DE/rand/1/bin powered by the cPRNG using both chaotic maps. For each case the null hypothesis has been formulated according to the results mentioned in the Tables \ref{tab:DEBest1BinD10T}, \ref{tab:DEBest1BinD10G}, \ref{tab:DEbest1BinD20T}, \ref{tab:DEbest1BinD20G}, \ref{tab:DEbest1BinD30T} and \ref{tab:DEbest1BinD30G} for DE/best/1/bin and \ref{tab:DErand1BinD10T}, \ref{tab:DErand1BinD20T},\ref{tab:DErand1BinD30T} for DE/rand/1/bin. The significance level has been chosen to be $\alpha=0.1$ (To save space DE/best/1/bin will be denoted as DE/best and DE/rand/1/bin as DE/rand): 

\begin{itemize}
	\item DE/best, Tink.: $H_{0}$: The mean $\mu_{A}$ of Atan2 is greater than the mean $\mu_{B}$ of Bounds and mean $\mu_{M}$ of Modulo. 
	\item DE/best, Ging.: $H_{0}$: The mean $\mu_{B}$ of Bounds is greater than the mean $\mu_{A}$ of Atan2 and mean $\mu_{M}$ of Modulo. 
	\item DE/rand, Tink.: $H_{0}$: The mean $\mu_{B}$ of Bounds is greater than the mean $\mu_{A}$ of Atan2 and mean $\mu_{M}$ of Modulo. 

\end{itemize}

The results of the T-tests for cPRNGs are mentioned in the Table \ref{tab:stat}. In the case where we have denied the null hypothesis that one tested normalization schemes has greater mean than the second one we have used the two-sided T-test to find out if the means can be considered as equal and in the column p-value we mention the value from the two-sided T-test (denoted by ($t_2$) ), else p-value from the one-sided T-test is mentioned. The last column denotes if the null hypothesis $H_{0}$ has been accepted (Acc.) or denied (Den.).

The goal of the second part of this paper was to find out if the fruitfulness of the cPRNG used in DE is affected just by its PDF or it also depends on the specific sequences of numbers generated by cPRNGs. From this reason we have modified the PDF of MT to generate numbers with the same distribution like our cPRNGs using given  chaotic map and particular number normalization Atan2, Bounds, and Modulo. Results are mentioned in the Tables \ref{tab:DEBest1BinD10T} -- \ref{tab:DErand1BinD30G} (even columns). We have verified by Kolmogorov-Smirnov test that the results mentioned in the columns of the Tables \ref{tab:DEBest1BinD10T} -- \ref{tab:DErand1BinD30G} are normally distributed. ANOVA could not be used in DE/best/1/bin powered by MT with the modified PDF according to the cPRNG using Tinkerbell (tMT), DE/rand/1/bin powered by tMT and DE/rand/1/bin with the modified PDF according to the cPRNG using Gingerbread man (gMT) because variances of data sets could not be consider as equal. In the case of DE/best/1/bin powered by gMT the variances could be considered as equal and the test ANOVA has been applied, see Table \ref{tab:Anova1}. Subsequently, we have formulated null and alternative hypothesis and T-test (one-sided as well as two-sided denoted as $t_2$) has been applied to the remaining data sets:

\begin{itemize}
	\item DE/best, tMT: $H_{0}$: The mean $\mu_{A}^{tMT}$ is greater than the mean $\mu_{B}^{tMT}$ and mean $\mu_{M}^{tMT}$. $H_{A}$: The mean $\mu_{A}^{tMT}$ is not greater than the mean $\mu_{B}^{tMT}$ and mean $\mu_{M}^{tMT}$. \textbf{Interpretation:} MT with modified PDF according to the cPRNG using Tinkerbell and normalization scheme Atan2 is the most successful from the view of DE/best convergence speed (in comparison with tMT using Bounds and Modulo).
	\item DE/best, gMT.: In this case it was found out by the test ANOVA that the means can be considered as equal, see Table \ref{tab:Anova1}. 
	\item DE/rand, tMT.: $H_{0}$: The mean $\mu_{M}^{tMT}$ is greater than the mean $\mu_{A}^{tMT}$ and the mean $\mu_{B}^{tMT}$. $H_{A}$: The mean $\mu_{M}^{tMT}$ is not greater than the mean $\mu_{A}^{tMT}$ and the mean $\mu_{B}^{tMT}$. \textbf{Interpretation:} MT with modified PDF according to the cPRNG using Tinkerbell and normalization scheme Modulo is the most successful from the view of DE/rand convergence speed (in comparison with tMT using Atan2 and Bounds). 
	\item DE/rand, gMT: $H_{0}$: The mean $\mu_{B}^{gMT}$ is greater than the mean $\mu_{A}^{gMT}$ and the mean $\mu_{M}^{gMT}$. $H_{A}$: The mean $\mu_{B}^{gMT}$ is not greater than the mean $\mu_{A}^{gMT}$ and the mean $\mu_{M}^{gMT}$. \textbf{Interpretation:} MT with modified PDF according to the cPRNG using Tinkerbell and normalization scheme Bounds is the most successful from the view of DE/rand convergence speed (in comparison with tMT using Atan2 and Modulo). 
\end{itemize}

The results are mentioned in the Table \ref{tab:statMT}. In the last column we explicitly indicate whether the null hypothesis $H_{0}$ has been accepted (Acc.) or denied (Den.). 

To find out whether the PDF of the cPRNG fundamentally affect DE convergence speed we have compared the results of cPRNGs (using number normalization Atan2, Bounds, Modulo)  and MT with modified PDF mentioned in the Tables \ref{tab:DEBest1BinD10T} -- \ref{tab:DErand1BinD30G}. We have compared the means of the best results if the cPRNG and modified MT reach the best results in the same category (Atan2, Bounds, Modulo). If all categories reach the comparable results we compare means of all categories ($\mu_{A}$ vs. $\mu_{A}^{MT}$, $\mu_{B}$ vs. $\mu_{B}^{MT}$ \ldots). If the number generators reach the best results in the different category, we compare these categories at the end of our work (in the case of DE/rand, Ging. vs gMT).  It was found out by the F-test that the variances of the comparing results can be considered as equal. We have formulated hypothesis and T-test has been used. The significance level has been chosen to be $\alpha=0.1$. The results are mentioned in the Table \ref{tab:statFin}.

\begin{itemize}
	\item DE/best, Tink vs. tMT, Atan2: $H_{0}$: The mean $\mu_{A}$ is equal to the mean $\mu_{A}^{tMT}$. \textbf{Interpretation:} The sequence of the numbers generated by the cPRNG does not significantly influence DE convergence speed.
	\item DE/best, Ging. vs. gMT, Atan2: $H_{0}$: The mean $\mu_{A}$ is equal to the mean $\mu_{A}^{gMT}$. \textbf{Interpretation:} The sequence of the numbers generated by the cPRNG does not significantly influence DE convergence speed.
	\item DE/best, Ging vs. gMT, Bounds: $H_{0}$: The mean $\mu_{B}$ is equal to the mean $\mu_{B}^{gMT}$. \textbf{Interpretation:} The sequence of the numbers generated by the cPRNG does not significantly influence DE convergence speed.
	\item DE/best, Ging vs. gMT, Bounds: $H_{0}$: The mean $\mu_{M}$ is equal to the mean $\mu_{M}^{gMT}$.\textbf{Interpretation:} The sequence of the numbers generated by the cPRNG does not significantly influence DE convergence speed.
	\item DE/rand, Tink. vs. gMT, Atan2: $H_{0}$: The mean $\mu_{A}$ is equal to the mean $\mu_{A}^{tMT}$. \textbf{Interpretation:} The sequence of the numbers generated by the cPRNG does not significantly influence DE convergence speed.
	\item DE/rand, Tink vs. gMT, Bounds: $H_{0}$: The mean $\mu_{B}$ is greater than the mean $\mu_{B}^{tMT}$. \textbf{Interpretation:} We can not say that the sequence of the numbers generated by the cPRNG does not significantly influence DE convergence speed.
	\item DE/rand, Tink vs. gMT, Bounds: $H_{0}$: The mean $\mu_{M}$ is smaller than the mean $\mu_{M}^{tMT}$. \textbf{Interpretation:} We can not say that the sequence of the numbers generated by the cPRNG does not significantly influence DE convergence speed.
	\item DE/rand, Tink. vs gMT, Bounds vs. Modulo: The mean $\mu_{B}$ is equal to the mean $\mu_{M}^{tMT}$.	\textbf{Interpretation:} We can not say that the sequence of the numbers generated by the cPRNG does not significantly influence DE convergence speed.
	\item DE/rand, Ging. vs. gMT: $H_{0}$: The mean $\mu_{A}$ is equal to the mean $\mu_{A}^{gMT}$. \textbf{Interpretation:} The sequence of the numbers generated by the cPRNG does not significantly influence DE convergence speed.
	\item DE/rand, Ging. vs. gMT: $H_{0}$: The mean $\mu_{B}$ is equal to the mean $\mu_{B}^{gMT}$. \textbf{Interpretation:} The sequence of the numbers generated by the cPRNG does not significantly influence DE convergence speed.
\end{itemize}

 Setting of DE ($D$ denotes dimension, $NP$ number of individuals, $F$ mutation constant, and $CR$ crossover probability):

\begin{itemize}
	\item $D=10$: $NP = 50$, $G=200$
	\item $D=20$: $NP = 100$, $G=400$
	\item $D=30$: $NP = 150$, $G=600$
	\item For all experiments $F=0.5$, $CR=0.85$
\end{itemize}

\section{Results}
\label{sec:res}

The results for DE/best/1/bin and DE/rand/1/bin with dimensions $D=10$, $D=20$, and $D=30$ are mentioned in the Tables \ref{tab:DEBest1BinD10T} -- \ref{tab:DErand1BinD30G}. The results describe the percentage of experiments, where the selected number normalization scheme applied to the number generated by the cPRNG has been the most successful. In other words in the tables mentioned above we can find the relative number of winnings of the normalization scheme using in the selected cPRNG (from the view of DE convergence speed). For the statistical analysis, please see \cite{web1}. The best results for DE using cPRNGs are marked in bold and the best results for DE using MT with the modified PDF are mentioned in the brackets. MT with the modified PDF according to the cPRNG using Gingerbread man map is denoted as gMT and Tinkerbell map as tMT.


\begin{table}[h!]
\renewcommand{\arraystretch}{1.2}
	\centering
		\caption{DE/best/1/bin, $D=10$, Tinkerbell (T) vs. tMT.}
		\begin{tabular}{l|rr|rr|rr}
		\hline
					&	\multicolumn{2}{c|}{\textbf{Atan2}}	&	\multicolumn{2}{c|}{\textbf{Bounds}}	&	\multicolumn{2}{c}{\textbf{Modulo}}\\
		 \cline{2-7}
			& \multicolumn{1}{c}{\textbf{T}} &	\multicolumn{1}{c|}{\textbf{tMT}} &	\multicolumn{1}{c}{\textbf{T}} &	\multicolumn{1}{c|}{\textbf{tMT}} &	\multicolumn{1}{c}{\textbf{T}} &	\multicolumn{1}{c}{\textbf{tMT}}\\
			\hline
		$f_1$ &	\textbf{74\%} &	(66\%) &	20\% &	34\% &	6\% &	0\% \\
		$f_5$ &	\textbf{90\%} &	(88\%) &	4\% &	12\% &	6\% &	0\%  \\
		$f_9$ &	0\% &	0\% &	\textbf{82\%} &	8\% &	22\% &	(94\%)\\
		$f_{13}$ &	22\% &	28\% &	36\% &	38\% &	\textbf{44\%} &	(40\%) \\
		$f_{15}$ &	30\% &	12\% &	\textbf{40\%} &	(44\%) &	30\% &	(44\%)\\
		$f_{16}$ &	2\% &	0\% &	\textbf{68\%} &	12\% &	40\% &	(88\%)\\
		$f_{17}$ &	\textbf{70\%} &	(46\%) &	14\% &	(46\%) &	16\% &	8\%\\
		$f_{22}$ &	\textbf{84\%} &	(66\%) &	14\% &	34\% &	2\% &	0\%\\
		$f_{23}$ &	\textbf{72\%} &	(58\%) &	16\% &	42\% &	12\% &	0\%\\
			\hline
		\end{tabular}
	\label{tab:DEBest1BinD10T}
\end{table}

\begin{table}[h!]
\renewcommand{\arraystretch}{1.2}
	\centering
		\caption{DE/best/1/bin, $D=10$, Gingerbread man (G) vs. gMT.}
		\begin{tabular}{l|rr|rr|rr}
			\hline
					&	\multicolumn{2}{c|}{\textbf{Atan2}}	&	\multicolumn{2}{c|}{\textbf{Bounds}}	&	\multicolumn{2}{c}{\textbf{Modulo}}\\
		 \cline{2-7}
			& \multicolumn{1}{c}{\textbf{T}} &	\multicolumn{1}{c|}{\textbf{tMT}} &	\multicolumn{1}{c}{\textbf{T}} &	\multicolumn{1}{c|}{\textbf{tMT}} &	\multicolumn{1}{c}{\textbf{T}} &	\multicolumn{1}{c}{\textbf{tMT}}\\
			\hline
		$f_1$ & \textbf{68\%} &	(50\%) &	0\% &	0\% &	32\% &	(50\%)\\
		$f_5$ &	\textbf{92\%} &	(100\%) &	0\% &	0\% &	8\% &	0\%\\
		$f_9$ &	 0\% &	0\% &	\textbf{100\%} &	(100\%) &	0\% &	0\%\\
		$f_{13}$ & 10\% &	14\% &	\textbf{64\%} &	(50\%) &	28\% &	42\%\\
		$f_{15}$ & 6\% &	6\% &	\textbf{70\%} &	(76\%) &	24\% &	18\%\\
		$f_{16}$ &	0\% &	0\% &	\textbf{96\%} &	(94\%) &	4\% &	6\%\\
		$f_{17}$ &	\textbf{58\%} &	(64\%) &	4\% &	14\% &	38\% &	22\%\\
		$f_{22}$ &	\textbf{96\%} &	(60\%) &	0\% &	6\% &	4\% &	34\%\\
		$f_{23}$ &	\textbf{92\%} &	(68\%) &	0\% &	4\% &	8\% &	28\%\\
			\hline
		\end{tabular}
	\label{tab:DEBest1BinD10G}
\end{table}

\begin{table}[h!]
\renewcommand{\arraystretch}{1.2}
	\centering
		\caption{DE/rand/1/bin, $D=10$, Tinkerbell (T) vs. tMT.}
		\begin{tabular}{l|rr|rr|rr}
		\hline
					&	\multicolumn{2}{c|}{\textbf{Atan2}}	&	\multicolumn{2}{c|}{\textbf{Bounds}}	&	\multicolumn{2}{c}{\textbf{Modulo}}\\
		 \cline{2-7}
			& \multicolumn{1}{c}{\textbf{T}} &	\multicolumn{1}{c|}{\textbf{tMT}} &	\multicolumn{1}{c}{\textbf{T}} &	\multicolumn{1}{c|}{\textbf{tMT}} &	\multicolumn{1}{c}{\textbf{T}} &	\multicolumn{1}{c}{\textbf{tMT}}\\
			\hline
			$f_{1}$ 	&	32\% &	32\% 	&	\textbf{44\%} &	(64\%) 	&	24\% &	14\% 	\\
			$f_{5}$ 	&	16\% &	20\% 	&	44\% &	(54\%) 	&	\textbf{46\%} &	36\% \\
			$f_{9}$ 	&	0\% &	2\% 		&	\textbf{70\%} &	20\% 	&	32\% &	(78\%) \\
			$f_{13}$ 	& 36\% &	26\% 	&	\textbf{40\%} &	26\% 	&	26\% &	(50\%) \\
			$f_{15}$ 	& 2\% &	12\% 	&	38\% &	22\% 	&	\textbf{60\%} &	(66\%) \\	
			$f_{16}$ 	& 4\% &	8\% 		&	\textbf{64\%} &	8\% 		&	34\% &	(84\%)\\
			$f_{17}$	& \textbf{40\%} &	(50\%) 	&	30\% &	30\% 	&	30\% &	20\%\\
			$f_{22}$ 	&	26\% &	32\% 	&	\textbf{56\%} &	(64\%) 	&	18\% &	4\%\\
			$f_{23}$ 	&	\textbf{64\%} &	42\% 	&	24\% &	(56\%) 	&	12\% &	2\%\\
			\hline
		\end{tabular}
	\label{tab:DErand1BinD10T}
\end{table}

\begin{table}[h!]
\renewcommand{\arraystretch}{1.2}
	\centering
		\caption{DE/rand/1/bin, $D=10$,  Gingerbread man (G) vs. gMT.}
		\begin{tabular}{l|rr|rr|rr}
			\hline
					&	\multicolumn{2}{c|}{\textbf{Atan2}}	&	\multicolumn{2}{c|}{\textbf{Bounds}}	&	\multicolumn{2}{c}{\textbf{Modulo}}\\
		 \cline{2-7}
			& \multicolumn{1}{c}{\textbf{T}} &	\multicolumn{1}{c|}{\textbf{tMT}} &	\multicolumn{1}{c}{\textbf{T}} &	\multicolumn{1}{c|}{\textbf{tMT}} &	\multicolumn{1}{c}{\textbf{T}} &	\multicolumn{1}{c}{\textbf{tMT}}\\
			\hline
			$f_{1}$ 	&	\textbf{68\%} &	(64\%) 	&	14\% &	30\% 	&	22\% &	20\%\\
			$f_{5}$ 	&	 14\% &	10\% 	&	\textbf{72\%} &	(74\%) 	&	22\% &	20\%\\
			$f_{9}$ 	&		0\% &	0\% 		&	\textbf{88\%} &	(96\%) 	&	14\% &	4\%\\
			$f_{13}$ 	&  24\% &	30\% 	&	\textbf{48\%} &	(44\%) 	&	28\% &	26\%\\
			$f_{15}$ 	&  4\% &	8\% 		&	\textbf{92\%} &	(84\%) 	&	4\% & 8\%\\	
			$f_{16}$ 	&  4\% &	2\% 		&	\textbf{94\%} &	(88\%) 	&	2\% & 10\%\\
			$f_{17}$	&  \textbf{60\%} &	(76\%) 	&	16\% &	4\% 		&	24\% &	20\%\\
			$f_{22}$ 	&	 \textbf{92\%} &	(94\%) 	&	0\% &	2\% 		&	8\% &	4\%\\
			$f_{23}$ 	&	\textbf{96\%} &	(96\%) 	&	2\% &	2\% 		&	2\% &	2\%\\
			\hline
		\end{tabular}
	\label{tab:DErand1BinD10G}
\end{table}


\begin{table}[h!]
\renewcommand{\arraystretch}{1.2}
	\centering
		\caption{DE/best/1/bin, $D=20$, Tinkerbell (T) vs. tMT.}
		\begin{tabular}{l|rr|rr|rr}
			\hline
					&	\multicolumn{2}{c|}{\textbf{Atan2}}	&	\multicolumn{2}{c|}{\textbf{Bounds}}	&	\multicolumn{2}{c}{\textbf{Modulo}}\\
		 \cline{2-7}
			& \multicolumn{1}{c}{\textbf{T}} &	\multicolumn{1}{c|}{\textbf{tMT}} &	\multicolumn{1}{c}{\textbf{T}} &	\multicolumn{1}{c|}{\textbf{tMT}} &	\multicolumn{1}{c}{\textbf{T}} &	\multicolumn{1}{c}{\textbf{tMT}}\\
			\hline
			$f_1$	&	\textbf{100\%}	&	(90\%) &	0\%	&	10\%	&	0\%	&	0\%	\\
			$f_5$	&	\textbf{98\%}	&	(90\%) &	2\%	&	10\%	&	0\%	&	0\%	\\
			$f_9$	&	0\%	&	0\%&	\textbf{74\%}	&	2\%	&	28\%	&	(100\%)	\\
			$f_{13}$	&	18\%	&	32\%	&	38\%	&	34\%	&	\textbf{46\%}	&	(36\%)\\
			$f_{15}$	&	6\%	&	2\%	&	\textbf{50\%}	&	26\%	&	48\%	&	(72\%)\\
			$f_{16}$	&	2\%	&	2\%	&	\textbf{72\%}	&	8\%	&	28\%	&	(90\%)\\
			$f_{17}$	&	\textbf{88\%}	&	(70\%)	&	12\%	&	30\%	&	0\%	&	0\%	\\
			$f_{22}$	&	\textbf{78\%}	&	(50\%)	&	18\%	&	48\%	&	4\%	&	2\%	\\
			$f_{23}$	&	\textbf{88\%}	&	(54\%)	&	12\%	&	46\%	&	0\%	&	0\%	\\
	\hline
		\end{tabular}
		\label{tab:DEbest1BinD20T}
\end{table}

\begin{table}[h!]
\renewcommand{\arraystretch}{1.2}
	\centering
		\caption{DE/best/1/bin, $D=20$, Gingerbread man (G) vs. gMT.}
		\begin{tabular}{l|rr|rr|rr}
			\hline
					&	\multicolumn{2}{c|}{\textbf{Atan2}}	&	\multicolumn{2}{c|}{\textbf{Bounds}}	&	\multicolumn{2}{c}{\textbf{Modulo}}\\
		 \cline{2-7}
			& \multicolumn{1}{c}{\textbf{T}} &	\multicolumn{1}{c|}{\textbf{tMT}} &	\multicolumn{1}{c}{\textbf{T}} &	\multicolumn{1}{c|}{\textbf{tMT}} &	\multicolumn{1}{c}{\textbf{T}} &	\multicolumn{1}{c}{\textbf{tMT}}\\
			\hline
			$f_1$	&	\textbf{62\%}	&	(100\%)	&	0\%	&	0\% &	38\%	&	0\%\\
			$f_5$	&	\textbf{88\%}	&	(100\%)	&	0\%	&	0\%&	12\%	&	0\%\\
			$f_9$	&	0\%	&	0\%	&	\textbf{98\%}	& (100\%)	&	2\%	&	0\%\\
			$f_{13}$	&	22\%	&	18\%	&	\textbf{52\%}	&	(54\%)	&	26\%	&	30\%\\
			$f_{15}$	&	0\%	&	2\%	&	\textbf{80\%}	& (82\%)&	26\%	&	16\%\\
			$f_{16}$	&	4\%	&	4\%	&	\textbf{86\%}	&	(94\%)	&	10\%	&	2\%\\
			$f_{17}$	&	\textbf{68\%}	&	32\%	&	0\%	&	4\%	&	32\%	&	(64\%)\\
			$f_{22}$	&	\textbf{62\%}	&	20\%	&	6\%	&	6\%	&	32\%	&	(74\%)\\
			$f_{23}$	&	\textbf{88\%}	&	10\%	&	6\%	&	0\%	&	38\%	&	(90\%)\\
	\hline
		\end{tabular}
		\label{tab:DEbest1BinD20G}
\end{table}

\begin{table}[h!]
\renewcommand{\arraystretch}{1.2}
	\centering
		\caption{DE/rand/1/bin, $D=20$, Tinkerbell (T) vs. tMT.}
		\begin{tabular}{l|rr|rr|rr}
			\hline
					&	\multicolumn{2}{c|}{\textbf{Atan2}}	&	\multicolumn{2}{c|}{\textbf{Bounds}}	&	\multicolumn{2}{c}{\textbf{Modulo}}\\
		 \cline{2-7}
			& \multicolumn{1}{c}{\textbf{T}} &	\multicolumn{1}{c|}{\textbf{tMT}} &	\multicolumn{1}{c}{\textbf{T}} &	\multicolumn{1}{c|}{\textbf{tMT}} &	\multicolumn{1}{c}{\textbf{T}} &	\multicolumn{1}{c}{\textbf{tMT}}\\
			\hline
			$f_1$ &	0\% &	0\% &	\textbf{60\%} &	8\% &	42\% &	(92\%)\\
			$f_5$	& 0\% &	0\% &	\textbf{66\%} &	2\% &	38\% &	(98\%)\\
			$f_9$	& 0\% &	0\% &	6\% &	0\% &	\textbf{94\%}	& (100\%)\\
			$f_{13}$ &	32\% &	(40\%) &	32\% &	22\% &	\textbf{38\%} &	38\% \\
			$f_{15}$ &	4\% &	10\% &	\textbf{60\%} &	14\% &	38\% &	(76\%) \\
			$f_{16}$ &	6\% &	10\% &	\textbf{66\%} &	16\% &	28\% &	(74\%) \\
			$f_{17}$ &	30\% &	34\% &	\textbf{38\%} &	(36\%) &	32\% &	30\% \\
			$f_{22}$ &	\textbf{48\%} &	42\% &	40\% &	(50\%) &	12\% &	8\% \\
			$f_{23}$ &	36\% &	30\% &	\textbf{54\%} &	(70\%) &	10\% &	0\% \\
			\hline
		\end{tabular}
	\label{tab:DErand1BinD20T}
\end{table}

\begin{table}[h!]
\renewcommand{\arraystretch}{1.2}
	\centering
		\caption{DE/rand/1/bin, $D=20$, Gingerbread man (G) vs. gMT.}
		\begin{tabular}{l|rr|rr|rr}
			\hline
					&	\multicolumn{2}{c|}{\textbf{Atan2}}	&	\multicolumn{2}{c|}{\textbf{Bounds}}	&	\multicolumn{2}{c}{\textbf{Modulo}}\\
		 \cline{2-7}
			& \multicolumn{1}{c}{\textbf{T}} &	\multicolumn{1}{c|}{\textbf{tMT}} &	\multicolumn{1}{c}{\textbf{T}} &	\multicolumn{1}{c|}{\textbf{tMT}} &	\multicolumn{1}{c}{\textbf{T}} &	\multicolumn{1}{c}{\textbf{tMT}}\\
			\hline
			$f_1$ &	\textbf{76\%} &	(52\%) &	24\% &	48\% &	0\% &	2\%\\
			$f_5$	& 8\% &	2\% &	\textbf{92\%} &	(100\%) &	0\% &	0\%\\
			$f_9$	& 0\% &	0\% &	\textbf{94\%} &	(98\%) &	6\% &	2\%\\
			$f_{13}$ &	28\% &	24\% &	\textbf{36\%} &	36\% &	\textbf{36\%} &	(40\%)\\
			$f_{15}$ &	6\% &	10\% &	\textbf{74\%} &	(72\%) &	20\% &	18\%\\
			$f_{16}$ &	2\% &	8\% &	\textbf{84\%} &	(80\%) &	14\% &	12\%\\
			$f_{17}$ &	\textbf{68\%} &	82\% &	8\% &	8\% &	24\% &	10\%\\
			$f_{22}$ &	\textbf{66\%} &	(90\%) &	2\% &	2\% &	32\% &	8\%\\
			$f_{23}$ &	\textbf{68\%} &	(100\%) &	6\% &	0\% &	26\% &	0\%\\

			\hline
		\end{tabular}
	\label{tab:DErand1BinD20G}
\end{table}

\begin{table}[h!]
\renewcommand{\arraystretch}{1.2}
	\centering
		\caption{DE/best/1/bin, $D=30$, Tinkerbell (T) vs. tMT. }
		\begin{tabular}{l|rr|rr|rr}
			\hline
					&	\multicolumn{2}{c|}{\textbf{Atan2}}	&	\multicolumn{2}{c|}{\textbf{Bounds}}	&	\multicolumn{2}{c}{\textbf{Modulo}}\\
		 \cline{2-7}
			& \multicolumn{1}{c}{\textbf{T}} &	\multicolumn{1}{c|}{\textbf{tMT}} &	\multicolumn{1}{c}{\textbf{T}} &	\multicolumn{1}{c|}{\textbf{tMT}} &	\multicolumn{1}{c}{\textbf{T}} &	\multicolumn{1}{c}{\textbf{tMT}}\\
			\hline
				$f_1$	& \textbf{98\%} &	(96\%) &	2\% &	0\%	&0\% &	0\% \\
				$f_5$	& \textbf{98\%} &	(96\%) &	2\% &	0\% &	0\% &	0\% \\
				$f_9$	& 0\%	 & 0\% &	\textbf{54\%} &	0\% &	46\% &	(100\%) \\
				$f_{13}$ & 32\% &	24\% &	\textbf{42\%} &	32\% &	28\% &	(46\%)\\
				$f_{15}$ &	6\% &	2\% &	42\% &	12\% &	\textbf{54\%} &	(90\%)\\
				$f_{16}$ &	4\%	& 0\% &	\textbf{66\%} &	6\% &	32\% &	96\% \\
				$f_{17}$ &	96\% &	(88\%) &	4\% &	12\% &	\textbf{100\%} &	0\% \\
				$f_{22}$ &	\textbf{100\%} &	(68\%) &	0\%	 & 32\% &	48\% &	0\%	\\
				$f_{23}$ &	\textbf{98\%} &	(64\%) &	2\% &	36\% &	0\% &	0\%\\
	\hline
		\end{tabular}
		\label{tab:DEbest1BinD30T}
\end{table}

\begin{table}[h!]
\renewcommand{\arraystretch}{1.2}
	\centering
		\caption{DE/best/1/bin, $D=30$, Gingerbread man (G) vs. gMT. }
		\begin{tabular}{l|rr|rr|rr}
			\hline
					&	\multicolumn{2}{c|}{\textbf{Atan2}}	&	\multicolumn{2}{c|}{\textbf{Bounds}}	&	\multicolumn{2}{c}{\textbf{Modulo}}\\
		 \cline{2-7}
			& \multicolumn{1}{c}{\textbf{T}} &	\multicolumn{1}{c|}{\textbf{tMT}} &	\multicolumn{1}{c}{\textbf{T}} &	\multicolumn{1}{c|}{\textbf{tMT}} &	\multicolumn{1}{c}{\textbf{T}} &	\multicolumn{1}{c}{\textbf{tMT}}\\
			\hline
				$f_1$	& 26\% & (100\%) &	0\% & 0\%	& \textbf{74\%} &	0\%\\
				$f_5$	& \textbf{84\%} &	(100\%) &	0\% &	0\% &	16\% &	0\%\\
				$f_9$	& 0\% &	0\% &	\textbf{100\%} &	(100\%) &	0\% &	0\%\\
				$f_{13}$ & 	22\% &	14\% &	\textbf{48\%} &	(54\%) &	32\% &	36\%\\
				$f_{15}$ &	0\% &	4\% &	\textbf{94\%} &	(88\%) &	6\% &	8\%\\
				$f_{16}$ &	2\% &	6\% &	\textbf{94\%} &	(92\%) &	4\% &	2\%\\
				$f_{17}$ &	0\% &	26\% &	0\% &	2\% &	\textbf{100\%} &	(72\%)\\
				$f_{22}$ &	\textbf{50\%} &	12\% &	2\% &	2\% &	48\% &	(86\%)\\
				$f_{23}$ &	\textbf{98\%} &	18\% &	4\% &	4\% &	58\% &	(78\%)\\
	\hline
		\end{tabular}
		\label{tab:DEbest1BinD30G}
\end{table}

\begin{table}[h!]
\renewcommand{\arraystretch}{1.2}
	\centering
		\caption{DE/rand/1/bin, $D=30$, Tinkerbell (T) vs. tMT. }
		\begin{tabular}{l|rr|rr|rr}
			\hline
					&	\multicolumn{2}{c|}{\textbf{Atan2}}	&	\multicolumn{2}{c|}{\textbf{Bounds}}	&	\multicolumn{2}{c}{\textbf{Modulo}}\\
		 \cline{2-7}
			& \multicolumn{1}{c}{\textbf{T}} &	\multicolumn{1}{c|}{\textbf{tMT}} &	\multicolumn{1}{c}{\textbf{T}} &	\multicolumn{1}{c|}{\textbf{tMT}} &	\multicolumn{1}{c}{\textbf{T}} &	\multicolumn{1}{c}{\textbf{tMT}}\\
			\hline
				$f_1$ & 0\%	&  0\% & \textbf{52\%} & 	0\%	 & 48\% & 	(100\%) \\
				$f_5$ & 0\% & 	0\% & 	\textbf{68\%} & 	0\% & 	32\% & (100\%)\\
				$f_9$ & 0\% & 	0\% & \textbf{74\%} & 	0\% & 28\% & 	(100\%) \\
				$f_{13}$ & 22\%	& 18\% & 	\textbf{46\%} & 	32\%	&  34\% & (50\%) \\
				$f_{15}$ & 12\% & 18\% & 	\textbf{64\%} & 	24\% & 	24\% & 	(58\%) \\
				$f_{16}$ & 16\% & 8\% & 	\textbf{54\%} & 	28\% & 	30\% & 	(64\%) \\
				$f_{17}$ & 6\% & 34\% & 	\textbf{48\%} & 	28\% & 	46\% & 	(38\%) \\
				$f_{22}$ & \textbf{54\%} & 	(92\%) & 	0\% & 	2\% & 46\% & 	6\%\\
				$f_{23}$ & 44\% & 	(92\%) & 	2\% & 	2\% & \textbf{54\%} & 	6\%\\
			\hline
		\end{tabular}
	\label{tab:DErand1BinD30T}
\end{table}

\begin{table}[h!]
\renewcommand{\arraystretch}{1.2}
	\centering
		\caption{DE/rand/1/bin, $D=30$, Gingerbread man (G) gMT.}
		\begin{tabular}{l|rr|rr|rr}
			\hline
					&	\multicolumn{2}{c|}{\textbf{Atan2}}	&	\multicolumn{2}{c|}{\textbf{Bounds}}	&	\multicolumn{2}{c}{\textbf{Modulo}}\\
		 \cline{2-7}
			& \multicolumn{1}{c}{\textbf{T}} &	\multicolumn{1}{c|}{\textbf{tMT}} &	\multicolumn{1}{c}{\textbf{T}} &	\multicolumn{1}{c|}{\textbf{tMT}} &	\multicolumn{1}{c}{\textbf{T}} &	\multicolumn{1}{c}{\textbf{tMT}}\\
			\hline
				$f_1$ &  0\% & 	0\% & 	\textbf{100\%} & 	(100\%) & 	0\% & 	0\%\\
				$f_5$ &  0\% & 	0\% & \textbf{100\% }& 	(100\%) & 	0\% & 	0\%\\
				$f_9$ &  0\% & 	0\% & 	\textbf{100\%} & (100\%) & 0\% & 0\%\\
				$f_{13}$ &  \textbf{38\%} & 30\% & 	\textbf{38\%} & (38\%) & 	26\% & 	32\%\\
				$f_{15}$ & 14\% & 10\% & 	\textbf{84\%} & (70\%) & 	2\% & 	20\%\\
				$f_{16}$ &  16\%	& 18\% & 	\textbf{74\%} & (64\%) & 	10\% & 	18\%\\
				$f_{17}$ &  20\% & (76\%) & 	12\% & 	18\% & 	\textbf{68\%} & 	6\%\\
				$f_{22}$ & \textbf{54\%} & 	(92\%) & 	0\% & 	2\% & 46\% & 	6\%\\
				$f_{23}$ & 44\% & 	(92\%) & 	2\% & 	2\% & \textbf{54\%} & 	6\%\\
			\hline
		\end{tabular}
	\label{tab:DErand1BinD30G}
\end{table}

\begin{table}[h!]
\renewcommand{\arraystretch}{1.2}
\centering
\caption{Results of the test ANOVA. The significance level has been chosen to be $\alpha=0.1$. The highlighted values mean that in these cases the means an be considered as equal.}
      \begin{tabular}{l|r|r|r}
				\hline
						Algorithm & cPRNG & p-value & F crit.\\
				\hline
					DE/best/1/bin & Tink. & 0.002  & 2.372\\
												& Ging. & \textbf{0.293} & 2.372\\
				\hline
					DE/rand/1/bin & Tink. & 0.000 & 2.372\\
				\hline
					DE/best/1/bin & gMT &  \textbf{0.610}& 2.372\\		
				\hline
			\end{tabular}			
			\label{tab:Anova1}  	
\end{table}

\begin{table}[h!]
\renewcommand{\arraystretch}{1.2}
\centering
\caption{Results of the T-tests for cPRNGs. The significance level has been chosen to be $\alpha=0.1$. If the p-value is smaller than the significance level value, the null hypothesis is accepted in the case of one-sided T-test. In the case of the two-sided T-test the null hypothesis is accepted if the p-value is greater than the significance level value.}
		\begin{tabular}{l|l|c|c|c|c}
		\hline
		\multicolumn{1}{c|}{Algorithm} & \multicolumn{1}{c|}{cPRNG}  & \multicolumn{1}{c|}{$\mu_{A}$ vs. $\mu_{B}$} & \multicolumn{1}{c|}{$\mu_{A}$ vs. $\mu_{M}$} & \multicolumn{1}{c|}{$\mu_{B}$ vs. $\mu_{M}$} & \multicolumn{1}{c}{Resume}\\
							&				 &	 \multicolumn{1}{c|}{(p-val.)}	&	 \multicolumn{1}{c|}{(p-val.)}	& \multicolumn{1}{c|}{(p-val.)}  &\\						
		\hline
		DE/best & Tink. &  $\mu_{A}>\mu_{B}$  & $\mu_{A}>\mu_{M}$   & $\mu_{B} =\mu_{M}$   & \textbf{Acc.}\\
						&				&  (0.006)					  &(0.001)              & (0.440)              &							\\
						\hline
		DE/rand & Tink. &  $\mu_{A}< \mu_{B}$ & $\mu_{A}>\mu_{M}$   & $\mu_{B} > \mu_{M}$  & \textbf{Acc.}\\
						&       & (0.000)             & (0.005)             & (0.000) 						 &							\\
						\hline
		DE/rand & Ging. &  $\mu_{A}< \mu_{B}$ & $\mu_{A}>\mu_{M}$   & $\mu_{B} > \mu_{M}$  & \textbf{Acc.}\\
						&				& (0.035)             & (0.025)             & (0.000)              &							\\
		\hline
		\end{tabular}
	\label{tab:stat}
\centering
\caption{Results of the T-tests for MT with modified PDF. The significance level has been chosen to be $\alpha=0.1$. If the p-value is smaller than the significance level value, the null hypothesis is accepted in the case of one-sided T-test. In the case of the two-sided T-test the null hypothesis is accepted if the p-value is greater than the significance level value.}
		\begin{tabular}{l|l|c|c|c|c}
		\hline
		\multicolumn{1}{c|}{Algorithm} & \multicolumn{1}{c|}{cPRNG}  & \multicolumn{1}{c|}{$\mu_{A}$ vs. $\mu_{B}$} & \multicolumn{1}{c|}{$\mu_{A}$ vs. $\mu_{M}$} & \multicolumn{1}{c|}{$\mu_{B}$ vs. $\mu_{M}$} & \multicolumn{1}{c}{Resume}\\
							&				 &	 \multicolumn{1}{c|}{(p-val.)}	&	 \multicolumn{1}{c|}{(p-val.)}	& \multicolumn{1}{c|}{(p-val.)}  &\\		
		\hline
		DE/best & tMT &  $\mu_{A}^{MT}>\mu_{B}^{MT}$  & $\mu_{A}^{MT}>\mu_{M}^{MT}$   & $\mu_{B}^{MT}=\mu_{M}^{MT}$  & \textbf{Acc.}\\
						&			& (0.003)												& (0.158)												&  (0.209)										 &\\
						\hline
		DE/rand & tMT &  $\mu_{A}^{MT}< \mu_{B}^{MT}$  & $\mu_{A}^{MT}<\mu_{M}^{MT}$  & $\mu_{B}^{MT} < \mu_{M}^{MT}$  & \textbf{Acc.}\\
						&			& (0.056)												 & (0.000)											& (0.005)												 &\\
						\hline
		DE/rand & gMT &  $\mu_{A}^{MT}= \mu_{B}^{MT}$   & $\mu_{A}^{MT}>\mu_{M}^{MT}$  & $\mu_{B}^{MT} > \mu_{M}^{MT}$ & \textbf{Acc.}\\
						&     & (0.304)													& (0.001)											 & (0.000) 											 &\\
		\hline
		\end{tabular}	
	\label{tab:statMT}
\end{table}

\begin{table}[h!]
\renewcommand{\arraystretch}{1.3}
	\centering
	\caption{The comparison of the best results of the cPRNGs and MT with the modified PDF. The significance level has been chosen to be $\alpha=0.1$. F crit. value equals to 1.675 in the one-sided test and 1.298 in the two-sided test ($t_2$). If the p-value is smaller than the significance level value, the null hypothesis is accepted in the case of one-sided T-test. In the case of the two-sided T-test the null hypothesis is accepted if the p-value is greater than the significance level value.}
		\begin{tabular}{l|l|l|l|r}
		\hline
		\multicolumn{1}{c|}{Algorithm}  & \multicolumn{1}{c|}{Comparison (norm. scheme)}  & \multicolumn{1}{c|}{Null hyp.} &  \multicolumn{1}{c|}{p-val.} & \multicolumn{1}{c}{Resume}\\
		\hline
		DE/best & Tink. vs. tMT (Atan2) & $\mu_{A}=\mu_{A}^{tMT}$ & 0.361 ($t_2$)  & \textbf{Acc.}\\
		\hline
						& Ging. vs. gMT (Atan2) & $\mu_{A}=\mu_{A}^{gMT}$& 0.544 ($t_2$) & \textbf{Acc.}\\
		DE/best & Ging. vs. gMT (Bounds) & $\mu_{B}=\mu_{B}^{gMT}$&0.944 ($t_2$) & \textbf{Acc.}\\
						& Ging. vs. gMT (Modulo) & $\mu_{M}=\mu_{M}^{gMT}$&0.776 ($t_2$) & \textbf{Acc.}\\
		\hline
						& Tink. vs. tMT (Atan2) & $\mu_{A}=\mu_{A}^{tMT}$& 0.530 ($t_2$) & \textbf{Acc.}\\ 
		DE/rand & Tink. vs. tMT (Bounds) & $\mu_{B}>\mu_{B}^{tMT}$& 0.000  & \textbf{Acc.}\\
						& Tink. vs. tMT (Modulo) & $\mu_{M}<\mu_{M}^{tMT}$& 0.010  & \textbf{Acc.}\\
						& Tink. vs. tMT (Bounds/Modulo) & $\mu_{B}=\mu_{M}^{tMT}$& 0.809 ($t_2$) & \textbf{Acc.}\\
		\hline
		DE/rand & Ging. vs gMT (Atan2) & $\mu_{A} = \mu_{A}^{gMT}$ & 0.455  ($t_2$) & \textbf{Acc.}\\
		DE/rand & Ging. vs. gMT (Bounds)  & $\mu_{B}=\mu_{B}^{gMT}$ & 0.983 ($t_2$) & \textbf{Acc.}\\
		\hline
		\end{tabular}	
	\label{tab:statFin}
\end{table}

\newpage
\section{Conclusion}
\label{sec:concl}

In this paper, we have been dealing with the effect of the normalization of the number generated by the chaotic maps to the DE convergence speed. Two chaotic maps -- Gingerbread man and Tinkerbell and two types of DE -- DE/best/1/bin and DE/rand/1/bin have been used. As the number normalization methods operation modulo (Modulo), straightforward number normalization where we know minimal and maximal generated number (Bounds) and arctangent (Atan2) of the two variables $x$ and $y$, where numbers $x$ and $y$ are outputs of the Gingerbread man map and Tinkerbell map have been chosen. Two first number normalization schemes have been successfully used in many publications mentioned above. The third scheme has been added to our work because there is no distortion of the PDF like in the case of operation modulo and we did not find the publication, where this scheme is used in this context. The goal of the first experiment was to find out if the normalization scheme used in the cPRNG which may generate numbers outside the unit interval can affect the DE convergence speed and which scheme is the most successful. In the second experiment, we have investigated the effect of the PDF of the cPRNG to the DE convergence speed. The main question was if the convergence speed of DE is influenced just by the PDF of the selected cPRNG or the numbers sequence plays a significant role in this process. From this reason we have modified MT to generate numbers with the same PDF like our cPRNGs using three schemes of the number normalization described above.

In the first experiments we have applied three schemes of number normalization to the numbers generated by Gingerbread man and Tinkerbell map. We have recorded how fast DE using these cPRNGs reaches the best results. If two schemes reach the best results in the same time, they are recorded as the best both. The results are mentioned in the Tables \ref{tab:DEBest1BinD10T} -- \ref{tab:DErand1BinD30G}. In the second experiment we have modified MT to generate numbers with the same probability like our cPRNGs. The results are mentioned in the same tables like in the case of the first experiment. 

As the most successful normalization scheme the scheme with the greatest mean is considered. Based on the results mentioned in the Tables \ref{tab:DEBest1BinD10T} -- \ref{tab:DErand1BinD30G} we have verified the normal distribution by Kolmogorov-Smirnov test. Then for each data sets -- DE/best/1/bin using Tinkerbell, DE/best/1/bin using Gingerbread man etc. we have tested if the variances can be considered as equal. In three data sets where MT with modified PDF had been used (DE/best/1/bin tMT, DE/rand/1/bin tMT and DE/rand/1/bin gMT) the variances could not be considered as equal. In the rest of data sets the statistical test ANOVA has been applied to find out if the means of normalization schemes can be considered as equal. The results mentioned in the Table \ref{tab:Anova1} showed the means of the categories (Atan2, Bounds and Modulo) of the DE/best/1/bin powered by Gingerbread man map and the means of the three categories (Atan2, Bounds and Modulo) of the DE/best/1/bin powered by gMT can be considered as equal.

For data sets where we have found out that the means of categories can not be considered as equal the statistical T-test has been used to find out which normalization scheme has been the most successful. The results are mentioned in the Tables \ref{tab:stat} and \ref{tab:statMT}. In the case of cPRNGs for DE/best/1/bin using Tinkerbell the normalization scheme \textbf{Atan2} has been the most successful. In the case of DE/rand/1/bin using Tinkerbell and DE/rand/1/bin using Gingerbread man the normalization scheme \textbf{Bounds} has been the most successful. In the case of DE/best/1/bin powered by tMT the normalization scheme \textbf{Atan2} has been the most successful. In the case of DE/rand/1/bin powered by tMT the normalization scheme \textbf{Modulo} has been the most successful and in the case of DE/rand/1/bin powered by gMT with the normalization schemes \textbf{Atan2} and \textbf{Bounds} has been the most successful. 

The last step of our work was to compare the results of the DE powered by cPRNGs using different number normalization and DE powered by MT with modified PDF according to these cPRNGs. When we look at the Table \ref{tab:statFin} we can make some conclusions:

\begin{itemize}
	\item DE/best/1/bin, Tink. vs. tMT, Atan2:  The means can be considered as equal. There is not significant difference between the cPRNG using Tinkerbell and number normalization Atan2 and PRNG MT with the modified PDF according to this cPRNG. \textbf{Interpretation:} The sequence of the numbers generated by this cPRNG does not significantly influence DE convergence speed.
	\item DE/best/1/bin, Ging. vs. gMT, Atan2: The means can be considered as equal. There is not significant difference between the cPRNG using Gingerbread man and number normalization Atan2 and PRNG MT with the modified PDF according to this cPRNG. \textbf{Interpretation:} The sequence of the numbers generated by this cPRNG does not significantly influence DE convergence speed.
	\item DE/best/1/bin, Ging. vs. gMT, Bounds: The means can be considered as equal. There is not significant difference between the cPRNG using Gingerbread man and number normalization Bounds and PRNG MT with the modified PDF according to this cPRNG. \textbf{Interpretation:} The sequence of the numbers generated by this cPRNG does not significantly influence DE convergence speed.
	\item DE/best/1/bin, Ging. vs. gMT, Modulo: The means can be considered as equal. There is not significant difference between the cPRNG using Gingerbread man and number normalization Modulo and PRNG MT with the modified PDF according to this cPRNG. \textbf{Interpretation:} The sequence of the numbers generated by this cPRNG does not significantly influence DE convergence speed.
	\item DE/rand/1/bin, Tink. vs. tMT, Atan2: The means can be considered as equal. There is not significant difference between the cPRNG using Gingerbread man and number normalization Atan2 and PRNG MT with the modified PDF according to this cPRNG. \textbf{Interpretation:} The sequence of the numbers generated by this cPRNG does not significantly influence DE convergence speed.
	\item DE/rand/1/bin, Tink. vs. tMT, Bounds: The means can not be considered as equal. The cPRNG using Tinkerbell and number normalization Bounds has reached better results (its mean is greater) than MT with the modified PDF according to this cPRNG. \textbf{Interpretation:} We can not say that the sequence of the numbers generated by the cPRNG does not significantly influence DE convergence speed.
	\item DE/rand/1/bin, Tink. vs. tMT, Modulo: The means can not be considered as equal. MT with the modified PDF according to the cPRNG using Tinkerbell and Modulo has reached better results (the mean is greater) than this cPRNG. \textbf{Interpretation:} We can not say that the sequence of the numbers generated by the cPRNG does not significantly influence DE convergence speed.
	\item DE/rand/1/bin, Tink. vs. tMT, Bounds vs. Modulo: In this case we have decided to compare two best number generators. We have compared cPRNG using Tinkerbell and number normalization Bounds and MT with the modified PDF according to the cPRNG using Tinkerbell and Modulo. The means are comparable. There is not significant difference between the cPRNG using Tinkerbell and number normalization Bounds and MT with the modified PDF according to the cPRNG using Tinkerbell and Modulo. \textbf{Interpretation:} We can not say that the sequence of the numbers generated by the cPRNG does not significantly influence DE convergence speed.
	\item DE/rand/1/bin, Ging. vs. gMT, Atan2: The means can be considered as equal. There is not significant difference between the cPRNG using Gingerbread man and number normalization Atan2 and PRNG MT with the modified PDF according to this cPRNG. \textbf{Interpretation:} The sequence of the numbers generated by this cPRNG does not significantly influence DE convergence speed.
	\item DE/rand/1/bin, Ging. vs. gMT, Bounds: The means can be considered as equal. There is not significant difference between the cPRNG using Gingerbread man and number normalization Bounds and PRNG MT with the modified PDF according to this cPRNG. \textbf{Interpretation:} The sequence of the numbers generated by this cPRNG does not significantly influence DE convergence speed.
\end{itemize}

From the results mentioned in the Section \ref{sec:res} we can say that the number normalization scheme might influence DE convergence speed. In our experiments, number normalization denoted as Atan2 and Bounds reached the best results. In the second part of our work, we were interested in the influence of the PDF and number sequences of the cPRNG using in DE. From the results mentioned above we can see that in three cases from four we can consider the means of the best results of the cPRNG and MT with the modified PDF according to this cPRNG as the same. That means that there is not significant difference between results of the cPRNG using the certain scheme of number normalization and MT with the modified PDF according to this cPRNG. 

In the case of DE/rand/1/bin cPRNG using Tinkerbell and Bounds reached the best results and MT with the modified PDF according to the cPRNG using Tinkerbell reached the best results with number normalization Modulo. On the other hand the results of the cPRNG using Tinkerbell and Bounds reached the comparable results like MT with modified PDF according to the cPRNG using Tinkerbell and Modulo. 

On the base of the results mentioned in the section \ref{sec:res} we express our opinion that the main role of the success of the cPRNG using in DE plays its PDF and the sequence of the numbers generated by this cPRNG is of secondary importance.

\newpage
\bibliographystyle{acm}      
\bibliography{mybib}   

\end{document}